# Density Weighted Connectivity of Grass Pixels in Image Frames for Biomass Estimation

Ligang Zhang[a*], Brijesh Verma[a], David Stockwell[b], Sujan Chowdhury[a]

[a]*School of Engineering and Technology, Central Queensland University, Brisbane, Australia*
[b]*Department of Transport and Main Roads, Emerald, Queensland, Australia*

{l.zhang, b.verma, s.chowdhury2}@cqu.edu.au; david.r.stockwell@tmr.qld.gov.au

**Abstract:** Accurate estimation of the biomass of roadside grasses plays a significant role in applications such as fire-prone region identification. Current solutions heavily depend on field surveys, remote sensing measurements and image processing using reference markers, which often demand big investments of time, effort and cost. This paper proposes Density Weighted Connectivity of Grass Pixels (DWCGP) to automatically estimate grass biomass from roadside image data. The DWCGP calculates the length of continuously connected grass pixels along a vertical orientation in each image column, and then weights the length by the grass density in a surrounding region of the column. Grass pixels are classified using feedforward artificial neural networks and the dominant texture orientation at every pixel is computed using multi-orientation Gabor wavelet filter vote. Evaluations on a field survey dataset show that the DWCGP reduces Root-Mean-Square Error from 5.84 to 5.52 by additionally considering grass density on top of grass height. The DWCGP shows robustness to non-vertical grass stems and to changes of both Gabor filter parameters and surrounding region widths. It also has performance close to human observation and higher than eight baseline approaches, as well as promising results for classifying low vs. high fire risk and identifying fire-prone road regions.

## 1. Introduction

Biomass, which is typically defined as the over-dry mass of the above ground portion of a group of vegetation in forestry (Vazirabad & Karslioglu, 2011), is one of the most important parameters of roadside vegetation, such as grasses and trees. Automatic estimation of grass biomass can be useful in various real-world applications, including monitoring roadside grass growth conditions, enforcing effective roadside management, and evaluating road safety. One typical example regarding the use of biomass is to identify the level of fire risk due to the presence of high, dense and dry roadside grasses, which are often characterised by high biomass. From the perspective of the transport, roadside grasses of high biomass can potentially become a big fire threat to the safety of vehicles, particularly in remotely located rural regions. Enforcing regular and frequent checks on roadside grass conditions by humans in a large state road network is often a big burden for transport authorities in terms of labour, cost, and time investments. Thus, it is of great significance to develop systems that are capable of automatically estimating the biomass of roadside grasses and precisely identifying those roadside regions with high fire risk, whereby necessary actions can be carried out to prevent possible fire threats such as burning or cutting the grasses.

* Corresponding Author. Telephone: +61 0732951162, Fax: +61 0732951162.

A typical method of calculating biomass is to conduct field surveys, which often include destructive plant sampling within a sampling region and calculating the weight after over-drying them (C. Royo & Villegas, 2011). It is one of the most accurate ways for obtaining biomass. Obviously, this method is heavily dependent on human efforts and requires extensive time, labour and cost, as well as expertise and equipment support. More importantly, it is unsuitable for automatic processing of data from large-scale fields.

The vast majority of existing solutions to automatically estimating the above-ground biomass of vegetation have been investigated using remote sensing methods (Lu, et al., 2016). The basic assumption of remote sensing methods for biomass estimation is that the mass of biomass is proportional to the volume of the vegetation and accordingly existing methods mainly base the biomass estimation on the upper layer of the canopy. The remotely sensed data can be captured using different types of sensors mounted on airborne, space-borne or terrestrial platforms. Optical spectral sensors are one of the most common ways of acquiring remotely sensed data with various spatial, spectral, radiometric and temporal resolutions. Typical examples of optical measurements are Vegetation Indices (VIs) (Schaefer & Lamb, 2016), spectral bands (Sibanda, Mutanga, & Rouget, 2016) and spatial image texture (Lu, et al., 2016). However, it is often difficult to obtain high quality optical data in frequent cloud conditions, and the optical measurements are prone to be affected by variations in solar radiation. Not all vegetation indices are closely related with biomass. The widely used Synthetic Aperture Radar (SAR) (Santi, et al., 2017) and LIght Detection And Ranging (LIDAR) (Lei Zhang & Grift, 2012), (Andújar, et al., 2016) sensors offer a better tolerance to weather and light conditions and are capable of collecting three dimensional distribution of structures within vegetation. Thus, they allow precise analysis on the characteristics of vegetation including biomass. However, these sensors are largely dependent on satellite or airborne platforms in existing studies, which leads to high costs and low flexibility. To provide more economical and convenient ways for data collection, more recent advances have tended to adopt drone-based sensors (Tang & Shao, 2015), (Kachamba, Ørka, Gobakken, Eid, & Mwase, 2016), (Fan, et al., 2017) or ground-based sensors such as ultrasonic sensor (Moeckel, Safari, Reddersen, Fricke, & Wachendorf, 2017), (Chang, et al., 2017) and mobile laser scanner (Ryding, Williams, Smith, & Eichhorn, 2015),(Li, Li, Zhu, & Li, 2016). Similar to satellite or airborne data, data collected using drone-mounted sensors reflect predominantly the upper canopy layers. Ground-based sensors can capture the whole above-ground structure of vegetation and thus they are suitable for biomass estimation in both large-scale and site-specific field surveys.

Except for remote sensing methods, another relatively less investigated method for biomass estimation is to use ground-based digital image or video data captured using ordinary cameras. Compared with remotely sensed data, ground-based images or video are relatively easier to collect using everyday devices such as ordinary cameras, smart phones and tablets, and can be operated by general people without requiring specialized knowledge. For the purpose of this paper, our industry partner - Department of Transport and Main Roads (DTMR), Queensland, Australia collects roadside video data from main state roads in Queensland using vehicle-mounted cameras, thereby human are employed to visually assess roadside conditions, such as vegetation species, height, fuel load, and potential safety threats to roads. For real-world applications where only ground-based video data are available, it is crucially important to develop automatic systems capable of estimating biomass from video frames.

Estimating biomass from ground-based digital image or video data is still a seldom investigated field. Studies (Juan & Xin-yuan, 2009), (Sritarapipat, Rakwatin, & Kasetkasem, 2014) exploited the way of estimating the height of vegetation from ground-based digital images. The height was calculated by measuring the distance between reference markers, which were pre-set manually on different parts of vegetation. Thus, these methods cannot be directly used for automatic applications. In our previous work (Verma, Zhang, & Stockwell, 2017), we have proposed the Vertical Orientation Connectivity of Grass Pixels (VOCGP) approach to automatically predict roadside grass biomass based on the grass height in images. The VOCGP approach segments brown grass pixels using an Artificial Neural Network (ANN) classifier with color and texture features, and detects the dominant texture direction at every pixel by performing Gabor-based votes on local texture. It then obtains the length of continuously connected grass pixels along every image column and takes an average length as the predicted biomass. However, the approach estimates the biomass predominantly based on the grass height and has largely ignored the contribution of the grass density to the biomass. The grass density is also an important for determining the grass biomass.

To solve the drawbacks in existing methods, this paper proposes Density Weighted Connectivity of Grass Pixels (DWCGP) to automatically estimate the biomass of roadside grasses in ground-based images. The DWCGP extends the VOCGP approach by jointly considering both grass height and density in the estimation of biomass, and thus it is expected that the DWCGP can produce more accurate estimation results. The main novelties in this paper are (a) a novel concept for determining the grass pixel orientation, height, and density without using any reference object; and (b) a novel integrated framework based on grass region segmentation and vertical grass orientation for grass biomass calculation. To the best of our knowledge, this is one of the first attempts that estimate grass biomass on ground-based data using image processing techniques.

The original contributions of this paper are as follows:

a) A concept of DWCGP for estimating grass biomass based on local texture features in a sampling window is presented. The DWCGP measures both the grass height and density to quantify the fuel loads of grasses, leading to accurate prediction of the biomass.

b) An integrated framework for DWCGP calculation based on the results of grass vs. non-grass pixel classification and vertical vs. non-vertical orientation detection is presented. Because the framework does not require manually setting up reference makers, nor the availability of specified equipment rather than a digital camera, it can be directly applied into site-specific analysis in a large-scale field.

c) An evaluation of DWCGP is presented by conducting a large number of experiments and comparisons with ground truths of both objective biomass and subjective density of roadside grasses collected from field surveys. A comparative analysis to show the effectiveness of DWCGP in supporting fire-prone road identification is included.

The remainder of the paper is organized as follows. Section 2 discusses related work. Section 3 introduces the proposed DWCGP approach. Experimental results are presented in Section 4. Section 5 draws the conclusions.

## 2. Related Work

This section reviews prior work on grass region segmentation and grass height estimation. Although intensive works (Hamuda, Glavin, & Jones, 2016) have been reported on vegetation or crop analysis and scene understanding, only few studies have specifically focused on roadside grass analysis. Compared with grassland vegetation, roadside grasses often have a more visible profile of the whole structure (e.g. appearance, geometry and length of grass stems), which is particularly important for analysing tall grasses. By contrast, analysis of grassland vegetation is often restricted to the upper layer of grasses only.

### *2.1 Grass Region Segmentation*

Existing work relevant to grass region segmentation can be approximately divided into two groups, including visible and invisible feature approaches.

a) Visible feature approaches extract visual properties of vegetation such as shape, texture, geometry, structure and color in the visible spectrum to distinguish them from other objects such as sky, road and soil. They can be further divided into three groups: (1) approaches that extract features from a Region Of Interest (ROI) for object classification. Campbell et al. (Campbell, Thomas, & Troscianko, 1997) adopted a self-organizing feature map for object segmentation using color and Gabor texture, and a multi-layer perceptron for classifying 11 outdoor objects. In (Haibing, Shirong, & Chaoliang, 2014), the mean shift was used to segment an image into local regions, and pixels in each region were mapped into the learnt Scale-Invariant Feature Transform (SIFT) words to obtain a histogram for recognizing five objects. In (Harbas & Subasic, 2014), the ROI in video was first estimated by optical flow, and color and Continuous Wavelet Transform (CWT) based texture were then extracted from the ROI for vegetation classification. Motion was also utilized in (Nguyen, Kuhnert, Thamke, Schlemper, & Kuhnert, 2012) to measure the resistance of vegetation pixels. (2) Approaches that perform object segmentation in a region merging process. In (Blas, Agrawal, Sundaresan, & Konolige, 2008), similar regions in road images were merged based on texton-based histogram profiles which were generated using K-means clustering on a fusion of L, *a*, *b* and pixel intensity differences, yielding 79% rate on classifying synthetic texture. In (Bosch, Muñoz, & Freixenet, 2007), the co-occurrence matrix was combined with RGB, HLS, and L*ab* to segment objects, and pixels were grown by minimizing a global energy function which integrates region and boundary information. In (Ligang Zhang, Verma, & Stockwell, 2016), roadside vegetation segmentation was achieved by progressively merging low confident superpixels to their most similar neighbors based on color and texture features. (3) Approaches that accomplish vegetation segmentation using prediction models. In (Zafarifar & de With, 2008), the pixel intensity difference was used in conjunction with a 3D Gaussian model of YUV for building a grass segmentation model, yielding 91% accuracy on 62 images. In (Schepelmann, Hudson, Merat, & Quinn, 2010), four statistic measures were clustered for segmenting illuminated grasses from artificial obstacles, achieving 95% accuracy on 40 region samples. In (Ligang Zhang, Verma, & Stockwell, 2015), opponent color intensity and color moments were used in conjunction with an ANN for vegetation classification. The local binary patterns and gray-level co-occurrence matrix features were combined with majority voting over three classifiers, including ANN, K-Nearest Neighbors (KNN) and Support Vector Machine (SVM), for dense vs. sparse vegetation classification in (Chowdhury, Verma, & Stockwell, 2015).

Visible features for object segmentation can also be extracted from remotely sensed data. In (Hu, Chen, Pan, & Hao, 2016), edges and initial object regions were extracted from satellite landscape images and used to find optimal objects by analysing the relationship between edges and regions in a region-growing process. In (Alshehhi, Marpu, Woon, & Mura, 2017), roads and buildings were segmented from satellite images using a patch-based Convolutional Neural Network (CNN). A set of shape features of adjacent regions was adopted to refine the segmentation results. Work (Cheng & Han, 2016) presented a survey of existing object segmentation (or detection) methods from remotely sensed images, and categorized them into four groups: template matching-based, knowledge-based, object-based, and machine learning-based. Although remotely sensed data in existing work are primarily aerial or satellite images, there have been increasingly more studies that have used drone-based image data (Malek, Bazi, Alajlan, AlHichri, & Melgani, 2014) and more convenient mobile laser scanning data (Li, et al., 2016).

Although promising results have been achieved, existing visible feature approaches still face the challenge of choosing or designing suitable visible features that are robust to the varieties in complicated real-life environments and scene content. Most evaluations are restricted to small evaluation data, and some studies just focus on artificial data which is far from the capacity of simulating real-world situations.

b) Invisible feature approaches utilize the reflectance properties of vegetation in the invisible spectrum to recognize them. One of the most important features is Vegetation Index (VI), which indicates the differences between the spectral properties of vegetation and those of other objects on invisible spectrum wavelengths, such as green and near infrared. As an instance, a simple comparison between red and Near Infrared Ray (NIR) reflectance has shown high accuracy of detecting photosynthetic vegetation (Bradley, Unnikrishnan, & Bagnell, 2007). To enhance its robustness against environmental effects, the NIR was later extended to various versions, such as the Normalized Difference Vegetation Index (NDVI) (Bradley, et al., 2007), the Modification of NDVI (MNDVI) (Nguyen, Kuhnert, & Kuhnert, 2012b), and combination of NDVI and MNDVI (Nguyen, Kuhnert, Thamke, et al., 2012). To fully utilize merits of both visible and invisible features, they were also integrated for vegetation segmentation. In (Nguyen, Kuhnert, Jiang, Thamke, & Kuhnert, 2011), 3D scatter features extracted from LADAR data were fused with histograms of HSV to segment vegetation. In (Y. Kang, Yamaguchi, Naito, & Ninomiya, 2011), 20-D filter bank features extracted from L, *a*, *b* color and infrared channels were integrated in a hierarchical architecture for road object recognition. In (Nguyen, Kuhnert, & Kuhnert, 2012a) the opponent color and Gabor features were fused to calculate visual similarity between neighbouring pixels, which was used for growing vegetation pixels from initial seed pixels. The initial pixels were selected based on a fusion of NDVI and MNDVI features.

Compared with visible feature approaches, invisible feature approaches often have better robustness against changes of environmental conditions, particularly in extreme lighting conditions such as a dark environment. However, most existing invisible approaches still require specialized equipment to capture VI features and face the challenge of designing VIs reliable under complicated real-world conditions. Recent advances on using drone-based or ground-based sensors have greatly promoted the application of invisible features to supporting object segmentation from remotely sensed data.

*2.2 Grass Biomass Estimation*

Related work on grass biomass estimation can be roughly classified into three groups, including human inspection, remote sensing methods, and image processing algorithms.

a) Visual inspection is a common approach to estimate the vegetation biomass in real practice, and it requires human going to the grass field and visually comparing the actual grasses with established criteria (e.g., ruler) to obtain the height of grasses. Although the results are generally accurate, human inspection is often labour-intensive, time-consuming, and costly. It also requires a certain degree of knowledge about field surveys and may need access permission from private landowners or relevant authorities. For field surveys in public roadsides as the case in this paper, access permission from private landowners is not required. However, permission from the government transport authority is still needed.

b) The vast majority of existing work on automatic biomass estimation are based on remote sensing measurements. The remotely sensed data can be collected using satellite-based, airborne or terrestrial equipment such as optical spectral (Sibanda, et al., 2016), LIDAR (Vazirabad & Karslioglu, 2011), SAR (Santi, et al., 2017), and ultrasonic sensors (Moeckel, et al., 2017), (Chang, et al., 2017). VI is one of the earliest and most popular spectral measurements for biomass estimation. In (Payero, Neale, & Wright, 2004), 11 types of VIs were compared for predicting the height of grasses and alfalfa. The results showed that four of them have strong linear relationships with the plant height. Thus, it is recommended to select an appropriate type of VI for every particular type of vegetation. The LIDAR sensor is another popular way of remote sensing data collection and it has showed higher accuracy and more precise information about the canopy than ultrasonic sensor and VIs (Llorens, Gil, Llop, & Escolà, 2011). One popular LIDAR model in determining vegetation height is the Canopy Height Model (CHM) (St-Onge, Hu, & Vega, 2008),(Grenzdörffer, 2014), which obtains the height by calculating the difference between the Digital Surface Model (DSM) and the Digital Terrain Model (DTM). However, the CHM requires the generation of both DSM and DTM. To remove the dependence on the DTM, Yamamoto et al. (Yamamoto, et al., 2011) proposed the use of a top surface model that is nearly parallel to the DTM. The model achieved accuracy close to one meter in measuring the mean tree height. For a summary of work on measuring plant height from LIDAR data, readers are referred to (Ahamed, Tian, Zhang, & Ting, 2011). Multi-frequency SAR data have also been long adopted for biomass estimation. In (Santi, et al., 2017), SAR data was used in conjunction with airborne LIDAR data to predict the forest biomass using an ANN predictor. Although these methods have achieved promising results, they are largely dependent on satellite or airplane platforms. Recent advances tend to employ more convenient and economical drone-based or ground-based sensors such as ultrasonic sensor (Moeckel, et al., 2017), (Chang, et al., 2017) and handheld mobile laser scanner (Ryding, et al., 2015). These platforms have greatly facilitated an easy and cheap deployment of various sensing equipment in both site-specific and large-scale field surveys, which has opened up new opportunities for a wide use of remote sensing techniques for biomass estimation. It is noted that features extracted from different sensors have also been combined in existing studies to provide more accurate estimation results, such as the fusion of VIs and terrestrial laser data (Tilly, Hoffmeister, et al., 2015), and fusion of ultrasonic and spectral sensor data (Moeckel, et al., 2017). For surveys of existing remote sensing methods for biomass estimation, readers are referred to (Lu, 2006), (Lu, et al., 2016), (Galidaki, et al., 2017).

c) Only few studies have investigated using image processing techniques for measuring the height of plants from ground-based image data. In (Sritarapipat, et al., 2014), a method was presented for measuring the rice height from dynamically monitored figures of a rice field. The height is measured by matching the height of rice against the pre-known height of a bar that was initially installed in the field. Work (Juan & Xin-yuan, 2009), (Dianyuan & Chengduan, 2011) adopted a similar idea for obtaining the height of trees, which is calculated via a proportion transform of the coordinates of three pre-set makers on the tree. The extension to the system was presented in (Dianyuan, 2011), which used three marker points and a perspective transformation. Essentially, the principal idea of these approaches is to transfer the task of measuring the plant height to the task of locating pre-set markers using image processing algorithms. However, they have a strict requirement of field settings such as the height, location and angle of cameras, as well as a need of assistance to manually install reference markers. Similar to human inspection, these approaches are workable only for site-specific analysis and cannot be used for large-scale field analysis.

To combat with shortcomings in current work using ground-based image data, this paper introduces a novel DWCGP approach to automatically estimate the biomass of roadside grasses from image frames. The approach utilizes the connectivity of vertically oriented grass pixels along both vertical and horizontal directions to determine the biomass. It is a direct extension to our previous approach (Verma, et al., 2017) by weighting grass height with surrounding grass density to more effectively considering impact of both height and density. It is fully automatic and supports both small and large scale field tests. We further illustrate the effectiveness of the DWCGP approach in a practical task of fire risk identification on video data collected from a state road in Queensland, Australia.

## 3. DWCGP Approach

This section describes the problem formulation of grass biomass estimation using image processing techniques, and then introduces the framework of the proposed DWCGP approach.

*3.1 Motivations of the Proposed Approach*

As reviewed in Section 2.2, there is no directly related approach that utilizes image processing techniques for estimating the biomass of roadside grasses from ground-based image data. To accomplish the goal of automatic estimation, we opt to follow and stimulate the traditional method of monitoring and calculating the fuel load of grasses (tonnes/ha) in field surveys, which often includes three steps: 1) collecting grass stems from a sampling region, 2) counting the number of grass stems, and 3) obtaining the total over-dry weight. The second step of counting the number of stems may not be necessary when only the final weight is needed. For the purpose of this paper, we follow studies on measuring crop biomass (C. Royo & Villegas, 2011), (Conxita Royo, Nazco, & Villegas, 2014), (Soriano, Villegas, Aranzana, del Moral, & Royo, 2016), which recorded and used the number of stems in the calculation of the biomass. The biomass equals to the product of average dry weight per plant and the number of plants per unit area. The three steps can be mathematically expressed using the following formula:

$$F_l = \frac{1}{N_s}\sum_{\mathrm{j}=1}^{N_s} s_j \times u_j \quad (1)$$

where, $F_l$ is the fuel load of grasses within the sampling region, $N_s$ is the number of grass stems in the region, $s_j$ indicates the length of the $j^{th}$ stem, and $u_j$ stands for the fuel load unit (e.g. fuel load per centimetre of grasses) for the $j^{th}$ stem. The fuel load $F_l$ is averaged over all stems.

For an image of roadside grasses, assume $S \in R^{H \times W}$ be the image window which corresponds to the sampling grass region selected in field surveys, and $H$ and $W$ indicate the number of rows and columns respectively in $S$. Let $X_j = \{x_{1j}, x_{2j}, \dots, x_{ij}, \dots, x_{Hj}\}^T$ be the $j^{th}$ column vector of $S$, the target of automatic grass biomass estimation in $S$ is to find a projection solution $\Phi$ that is capable of transferring properties of image column vectors in $S$ into an estimated fuel load $F_e$:

$$F_e = \Phi(S) \quad (2)$$

$$\text{s.t. } \min\,(\mathrm{F}_e - \mathrm{F}_l) \quad (3)$$

Equation (3) enforces a constraint that the difference between the estimated fuel load and the physically quantified fuel load should be as minimal as possible.

To automatically estimate biomass from images in a similar way to manually measuring the physically quantified fuel load in field surveys, we make two assumptions: 1) grass stems can be analogously represented by columns of image pixels, and 2) the fuel load unit $u_j$ is equal for all stems of the same type of vegetation. Although the first assumption is not strictly correct in theory since grass stems do not always grow perfectly vertically in real-world situations, it is anticipated to represent an approximation of the vertical parts of stems and thus an indication of the grass height. Note that robust extraction of grass stems with different orientations is still a challenging task itself. The two assumptions greatly simplify the problem and provide a way to calculate the estimated fuel load in (2) based on column lengths of grass pixels in $S$:

$$F_e = \Phi(S) = \frac{1}{W}\sum_{j=1}^{W} l_j \times a \quad (4)$$

where, $W$ is the number of columns in $S$, $l_j$ is the length of grass pixels in the $j^{th}$ column $X_j$, and $a$ is a constant that enables a direct comparison between $l_j \times a$ and fuel load $s_j \times u_j$ in (1).

Equation (4) illustrates the basic concept of the proposed DWCGP approach that estimates the fuel load of grasses using image processing techniques in a way similar to the traditional method in field surveys. Specifically, the proposed approach calculates (4) in three consecutive steps, including measuring the length $l_j$ for the $j^{\text{th}}$ column, weighting the length by grass density $d_j$ in surrounding columns, and integrating the weighted lengths $\widehat{l_j}$ of all columns:

$$l_j = \Psi_1(X_j) \quad (5)$$

$$\widehat{l_j} = l_j \times d_j \quad (6)$$

$$\mathrm{F}_e = \Psi_2(\widehat{l_1}, \widehat{l_2}, \dots, \widehat{l_W}) \quad (7)$$

where, the $\Psi_1$ is a mapping function from image pixels in a column to a scalar representing the length of the grass stem in the column, and $\Psi_2$ is a mapping function from all weighted lengths to an estimated fuel load $\mathrm{F}_e$.

The proposed DWCGP is designed based on common knowledge in agriculture and forestry that the biomass yield is closely related with the plant height (Tilly, Aasen, & Bareth, 2015). However, our observation found that using solely the grass height will lead to incorrect biomass estimation in some practical conditions. If we take sparsely high grasses as an instance, we may get a relatively medium value (compared to sparsely low or densely high grasses) in the estimation of the grass height, but the actual density and biomass may be very low. This problem also exists in other similar conditions, such as low and high grasses co-existing in the sampling region. To solve the problem, the DWCGP also considers the grass density within a surrounding region relative to a specific column, which also has big impact on the grass biomass. This also agrees with the current practice of grassland curing in relevant government authorities (e.g. Country Fire Authority, Victoria, Australia: www.cfa.vic.gov.au/grass), which use both grass height and density as two major factors in human visual measurement of fuel loads of grasses.

To provide a solution that considers both grass height and density, the DWCGP calculates the connectivity of grass pixels along a vertical orientation in each image column and then weights it by the grass density in neighbouring columns. This is inspired by work (Rasmussen, 2004) which detected the primary direction of texture in a large neighbourhood by integrating the dominant texture orientation at image pixels. In (Verma, et al., 2017), we proposed an approach which takes the average length of all columns as an estimation of grass biomass. However, the approach relies on only height information along a vertical direction, but ignores density information along a horizontal direction. The proposed DWCGP also considers grass density in surrounding regions to provide more accurate prediction. For roadside grass images, it is observed that densely high grasses are normally characterised by long unbroken pixel connectivity across a certain number of continuous columns, while sparsely low grasses are often represented by short broken connectivity as shown in Fig. 1.

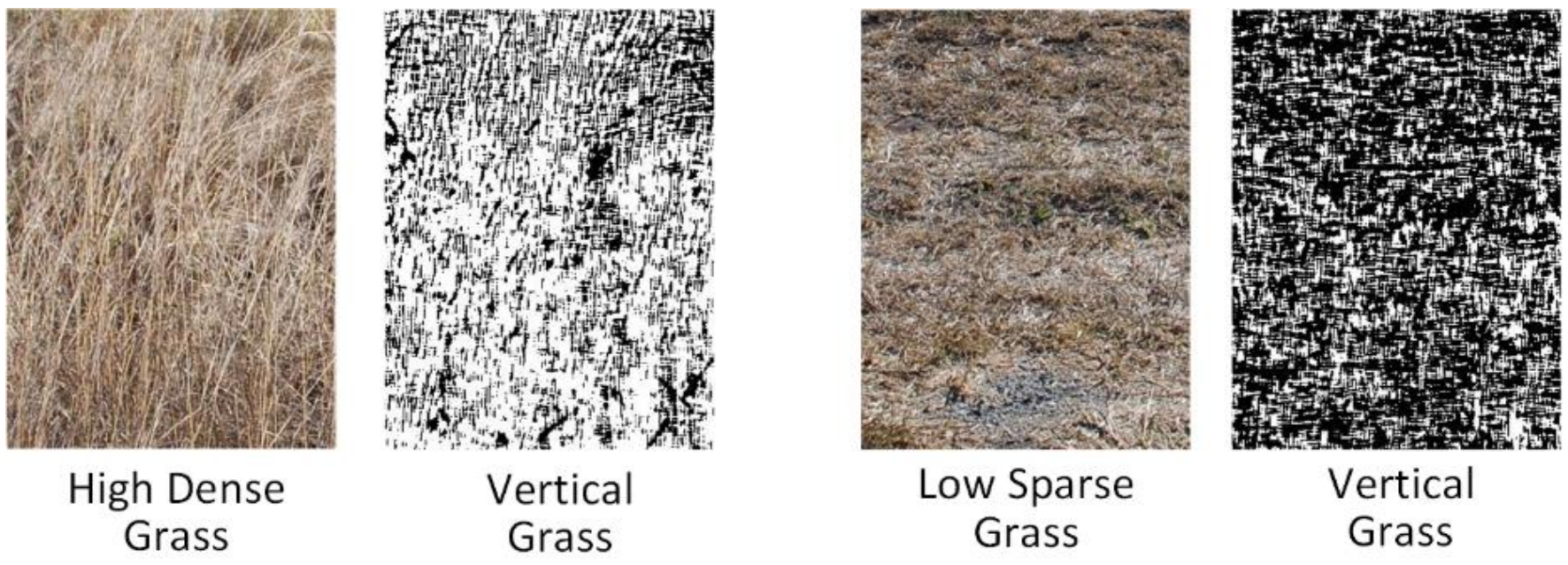

**Fig. 1.** Examples showing the differences in the classification results of grass pixels having a dominant vertical orientation between high dense grasses and low sparse grasses. Grass pixels with a vertical orientation are represented by a white color. Densely high grasses have longer connectivity along both vertical and horizontal orientations than sparely low grasses, which implies the importance of both height and density.

### *3.2 Framework of Proposed Approach*

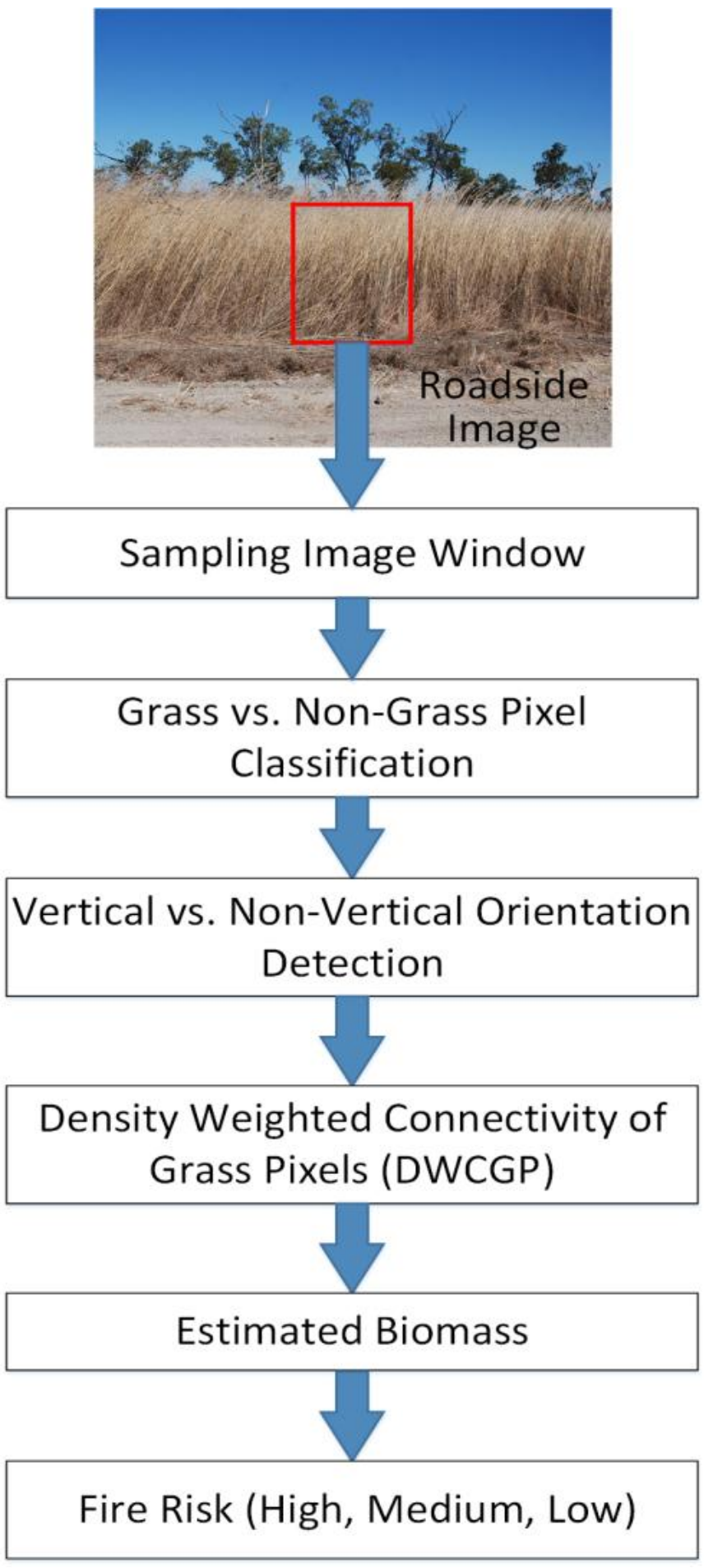

**Fig. 2.** Framework of the proposed DWCGP approach. For a sampling image window, the approach first classifies grass vs. non-grass pixels and then detects vertical vs. non-vertical dominant orientation at every pixel. The calculation of DWCGP is based on the connectivity of grass pixels along a vertical orientation in each column and the grass density in a surrounding region of the column.

Fig. 2 illustrates the systematic framework of the proposed DWCGP approach, which is composed of four main processing steps. The first step is to select a sampling grass window from a roadside image, which corresponds to the sampling grass region in field surveys, or can be any region of interest. The image window is used as the basic processing unit for biomass estimation. From the image window, grass region segmentation is performed to find the locations of all grass pixels using a feedforward ANN with color and texture features. As a parallel step, a Gabor filter vote process is employed to find the dominant texture orientation at each image pixel by performing a vote on the response magnitudes of multi-orientation and multi-scale Gabor filters on raw pixels in the grass window. Given the segmented grass pixels and their dominant orientations, a DWCGP algorithm is further presented to obtain the vertical connectivity of grass pixels in each image column and the grass density in a surrounding region, and then calculate average density weighted connectivity for all columns, which is taken as an estimated value of the biomass of grasses in the window.

### *3.3 Brown Grass Region Segmentation*

Brown grass region segmentation aims to label every pixel in the sampling window into brown grass vs. non-brown grass categories. We mainly focus on brown grasses as they often present a much larger fire threat than green grasses in practice. To effectively represent the visual characteristics of grass pixels, it is critical to select a set of suitable features. This paper adopts color and texture features as both of them convey important information for representing vegetation and other roadside objects.

For color features, we adopt six color channels from the CIEL*ab* and RGB spaces. The Lab channels are well-known for their high perceptually consistency with human vision, while RGB channels may provide complementary information to Lab channels. It is observed that brown grasses are predominantly represented by a yellow color, while other roadside objects such as sky, road and tree are primarily characterised by non-yellow colors.

For texture features, we employ features extracted using the 17-D filter banks which were first proposed in (Winn, Criminisi, & Minka, 2005). The 17-D filter banks were designed to generate a universal visual dictionary for object representation and they have shown high accuracy of real-life object recognition. Studies (Yousun Kang, Kidono, Naito, & Ninomiya, 2008) also indicated that they outperform Leung and Malik, Schmid, and MR8 filter sets for road object segmentation. The 17-D filter banks include Gaussians with three scales (1, 2, 4) applied to L, $a$, and $b$ channels, LoGs with four scales (1, 2, 4, 8) and the derivatives of Gaussians with two scales (2, 4) for each axis ($x$ and $y$) on the L channel.

The final 23-D color and texture feature set can be expressed using:

$$V_{i,j} = \{R, G, B, L, a, b, G^L_{1,2,4}, G^a_{1,2,4}, G^b_{1,2,4}, LOG^L_{1,2,4,8}, DOG^L_{2,4,x} DOG^L_{2,4,y}\} \quad (8)$$

where, $R, G, B, L, a, b$ represent color channels and $G^L_{1,2,4}, G^a_{1,2,4}, G^b_{1,2,4}, LOG^L_{1,2,4,8}, DOG^L_{2,4,x}$ $DOG^L_{2,4,y}$ represent texture features extracted using 17-D filter banks.

The extracted features can then be used for training a brown grass classifier. Amongst various types of classification algorithms, such as ANN, SVM, KNN and random forests, this paper chooses to use a three-layer feedforward ANN classifier due to its popularity and generalized capacity for classification. The ANN is composed of an input layer with 23 neurons, a hidden layer with $N$ neurons and an output layer with one neuron. Given the extracted color and texture features $V_{i,j}$, the input layer receives $V_{i,j}$ as input, performs multiplication and addition operations on $V_{i,j}$ based on a set of weights and constants, and finally transfers the calculated values using a linear or non-linear activation function:

$$\hat{V}_{i,j} = tran(wV_{i,j} + b) \tag{9}$$

where, $\hat{V}_{i,j}$ is the transferred output for $V_{i,j}$, $w$ and $b$ are trained weights and constants. $tran$ indicates the activation function, which is tangent sigmoid. The values $\hat{V}_{i,j}$ are again taken as input of neurons in the hidden layer and similar operations as in (9) are performed from the hidden layer to the output layer. Finally, the output layer produces a probability $p_{i,j}$ using a linear activation function, and $p_{i,j}$ indicates the likelihood of brown grasses for each pixel at coordinates $(i, j)$. The decision on the label of the pixel can then made based on $p_{i,j}$:

$$A_{i,j} \in \begin{cases} brown\ grass, & if\ \ p_{i,j} \geq T \\ no - brown\ grass, & others \end{cases} \tag{10}$$

where, $A_{i,j}$ represents a binary label for a pixel at ($i$, $j$), and $T = 0.5$ is a threshold. The output $A_{i,j}$ forms the foundation of calculating the connectivity and density of grass pixels.

*3.4 Vertical vs. Non-vertical Orientation Detection*

To determine grass height using the proposed approach, it is a prerequisite to accurately find the dominant texture orientation at every image pixel. This is achieved by performing a vote on Gabor filter responses along multi-orientations to determine the strongest texture orientation at each pixel. The use of Gabor filter is inspired by its excellent capacity in capturing multi-scale and multi-orientation texture such as line and edge, which are common characteristics of grass stems in images.

For the sampling window $S$, we first obtain its grey version by replacing each pixel by its average R, G and B values. Assume $F_{\theta,\phi}$ be 2D Gabor filter with an orientation $\theta$ and a scale $\phi$, the responses of $F_{\theta,\phi}$ from $S$ can be calculated by convolving the filter with intensities of all pixels in $S$:

$$G_{\theta,\phi} = S \oplus F_{\theta,\phi} \tag{11}$$

Because the output $G_{\theta,\phi}$ is a complex value, we combine its real and imaginary components using a square norm, yielding a complex magnitude:

$$\bar{G}_{\theta,\phi} = \sqrt{Real(G_{\theta,\phi})^2 + Img(G_{\theta,\phi})^2} \tag{12}$$

The $\bar{G}_{\theta,\phi}$ can be used as an indicator of the absolute strength of Gabor filter responses. However, since the proposed approach is primarily dependent on orientation information, we remove the scale information by keeping only the maximum response of all scales for each orientation:

$$\bar{G}_{\theta} = \max_{m=0,\ldots,M_{\phi}-1} \left(\bar{G}_{\theta,\phi_m}\right) \tag{13}$$

where, $M_{\phi}$ is the number of all scales.

Given a total of $N_\theta$ orientations, an orientation vector comprising of response magnitudes along all $N_\theta$ orientations can be obtained for every pixel at $(i, j)$:

$$\bar{G}_{i,j} = [\bar{G}_0^{i,j}, \bar{G}_1^{i,j}, \dots, \bar{G}_{N_\theta-1}^{i,j}] \tag{14}$$

The dominant orientation of every pixel at $(i, j)$ is then determined as the one having the maximum response among those of all orientations. This is essentially equivalent to performing a vote on response magnitudes along all orientations:

$$D_{i,j} = \tau \quad if \ \bar{G}_\tau^{i,j} = \max(\bar{G}_{i,j}) = \max_{k=0,\dots,N_\theta-1} \bar{G}_k^{i,j} \tag{15}$$

Because we consider only vertical vs. non-vertical orientations, $D_{i,j}$ is further converted into a binary category $O_{i,j}$ for a pixel at $(i, j)$:

$$O_{i,j} = \begin{cases} \text{vertical}, & if \ D_{i,j} = \upsilon \\ \text{non} - \text{vertical}, & others \end{cases} \tag{16}$$

where, $\upsilon$ indicates the index of the vertical orientation, and $0 \le \upsilon \le N_\theta - 1$.

*3.5 DWCGP Calculation*

Given the segmented grass vs. non-grass label and the detected vertical vs. non-vertical orientation for every image pixel, we further present an unsupervised DWCGP approach to predict the biomass in a sampling image window. The approach is composed of three major steps: a) determining the longest length of continuously connected grass pixels along a vertical orientation in every column; b) calculating the grass density in a surrounding region of the column, and c) obtaining density weighted lengths for each column and their average value over all columns. Thus, the DWCGP indicates an average of density weighted length of continuously connected grass pixels along a vertical orientation over all image columns and it is adopted as an estimation of the biomass.

Assume $A_{i,j} \in$ {grass; non-grass} be the segmented grass label and $O_{i,j} \in$ {vertical; non-vertical} be the detected dominant orientation for the pixel $x_{ij}$ at the $i^{th}$ row and $j^{th}$ column in the sampling window $S$, the value of $x_{ij}$ can be converted into a binary value as follows:

$$x_{ij} = \begin{cases} 1, \ if \ A_{i,j} = grass \text{ and } O_{i,j} = vertical \\ 0, \ others \end{cases} \tag{17}$$

The output $x_{ij}$ often contains some isolated pixels that are either non-grass or non-vertical, i.e., $x_{ij} = 0$, but are surrounded by vertical grass pixels. Those isolated pixels may severely impact the calculation of connectivity of grass pixels. To reduce the impact, those pixels are reassigned as vertical grass pixels as long as both its vertical neighbours are vertical grass pixels:

$$x_{ij} = 1 \ if \ x_{ij} = 0 \text{ and } x_{i+1,j} = 1 \text{ and } x_{i-1,j} = 1 \tag{18}$$

Note that boundary pixels in the region remain unchanged.

For all pixels $x_{ij}$, $1 \le i \le H$, in the $j^{th}$ column $X_j$ of $S$, the lengths of continuously connected pixels $x_{ij} = 1$ can be calculated:

$$C^j = \{c_1^j, c_2^j, \dots, c_Q^j\} \tag{19}$$

where, $c_q^j$ indicates the $q^{th}$ length in $X_j$, $Q$ is the total number of lengths in $X_j$, and $Q<=H$. Due to the existence of background noise and grass segmentation error, multiple lengths can be often obtained. To minimize the impact of those noise or error on the length results, only the longest length is kept and used as the length measurement $l_j$ for the $j^{th}$ column:

$$l_j = \max_{q=1,\dots,Q} c_q^j \tag{20}$$

To take into account grass density, let assume $R_j$ be a surrounding region of the $j^{th}$ column and having a width of $w_j$ pixels as shown in Fig. 3. The grass density within $R_j$ can be obtained by taking the percentage of grass pixels:

$$d_j = \frac{N_j}{w_j \times H} \tag{21}$$

where, $N_j$ indicates the number of grass pixels in $R_j$ and it can be calculated based on $A_{i,j}$, i.e., $A_{i,j} = grass$ and $A_{i,j} \in R_j$.

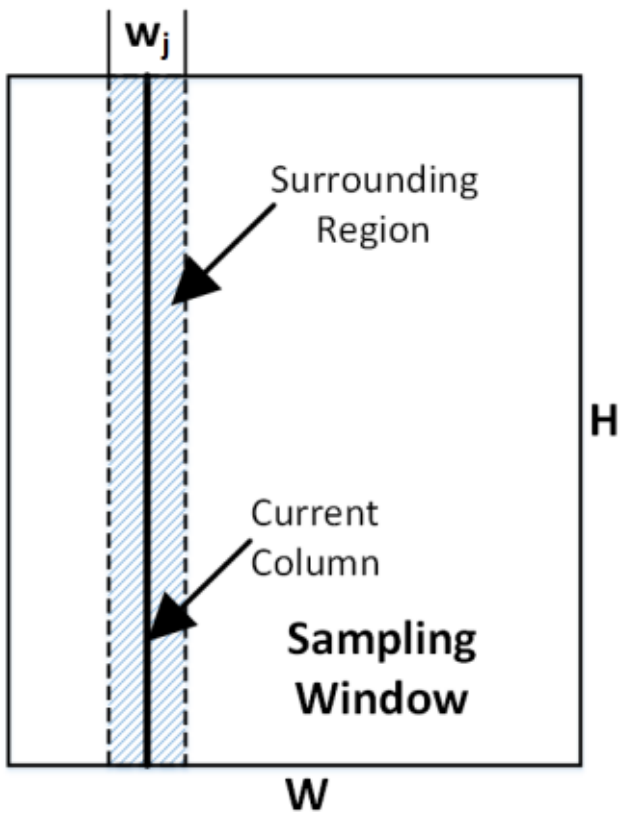

**Fig. 3.** Illustration of the surrounding region of a current column where the percentage of grass pixels is calculated. The percentage provides contextual information about grass density around every column.

We then can obtain the density weighted length for $j^{th}$ column based on (20) and (21):

$$\widehat{l_j} = l_j \times d_j \tag{22}$$

For all columns $X_j$ in $S$, $1 \le j \le W$, we can get all their length measurements using (23):

$$L = \{\widehat{l_1}, \widehat{l_2}, \dots, \widehat{l_W}\} \tag{23}$$

The final DWCGP in $S$ can be obtained by taking the average of density weighted length measurements of all columns:

$$\text{DWCGP} = \frac{1}{W}\sum_{j=1}^{W} \widehat{l_j} \tag{24}$$

One problem arising is that the DWCGP in (24) and the physically quantified fuel load in (1) are calculated based on different measurement units. To make them directly comparable, one possible solution is to set a calibration factor $a$ to the DWCGP:

$$\text{F}_e = a \times \text{DWCGP} \tag{25}$$

Until now, we have been able to use DWCGP as an estimated value of biomass. It should be noted that (20), (22) and (24) correspond respectively to (5), (6) and (7) in the problem formulation.

**Algorithm 1**: Pseudo-code of the DWCGP approach.

**Input:** Let $S$ be the sampling window of a height $H$ and a width $W$, $A_{i,j}$ ∈{grass; non-grass} and $O_{i,j}$ ∈{vertical; non-vertical} be labels of pixel $x_{ij}$ at coordinate ($i, j$), $1 \leq i \leq H, 1 \leq j \leq W$.
**Output:** DWCGP.

Initialize $L$ to empty
For $j^{th}$ column in $S$
    Initialize $C^j$ to empty
    Initialize $Length$ to zero
    For $i^{th}$ row in $S$
        If $A_{i,j}$ is *non-grass*
            If $Length$ is not equal to zero
                Add $Length$ to $C^j$
                Set $Length$ to zero
            End If
        Else
            If $O_{i,j}$ is *vertical*
                Add one to $Length$
                If $O_{i,j}$ is *last pixel in $j^{th}$ column*
                    Add $Length$ to $C^j$
                End If
            Else
                If $Length$ is not equal to zero
                    Add $Length$ to $C^j$
                    Set $Length$ to zero
                End If
        End If
    End For
    Find the longest length in $C^j$ using: $C^j = \{c_1^j, c_2^j, \dots, c_Q^j\}$ and $l_j = \max_{q=1,\dots,Q} c_q^j$
    Get grass density in surrounding region $R_j$: $d_j = N_j/(w_j \times H)$
    Get density weighted length: $\widehat{l_j} = l_j \times d_j$
    Add the weighted length $\widehat{l_j}$ to $L$: $L = \cup\ (L, \widehat{l_j})$
End For
Get DWCGP by averaging all elements in $L$ using: $\text{DWCGP} = \frac{1}{W}\sum_{j=1}^{W} \widehat{l_j}$
Get estimated biomass using: $\text{F}_e = a \times \text{DWCGP}$

The whole process of the proposed DWCGP approach is summarized in Algorithm 1. From the first to the last column in the sampling window, the approach begins by scanning the first pixel in every column and checking whether the pixel belongs to grass or non-grass (based on ANN classifier) and has a vertical or non-vertical orientation (based on Gabor filter votes). If the pixel is either non-grass or non-vertical, the approach continues moving to scan and check the next pixel in the column. Otherwise, if the pixel is grass and has a vertical orientation, it starts to count the number of continuously connected grass pixels having a vertical orientation along the same column. The counting continues until a non-grass or non-vertical orientation pixel is found, yielding multiple lengths of connected grass pixels for the column. However, we keep only the longest length in the column as it is a more robust indicator of the grass height than short lengths. For each column, we also calculate the percentage of grass pixels which is used as an indicator of the grass density of its surrounding region with a width of $w_j$ pixels. The longest length is then weighted with the density to take into account both grass height and density information. After obtaining the density weighted longest lengths of all columns, an 'average' operation is employed to obtain the DWCGP. This 'average' operation produces an average weighted height of all grasses in the sampling window.

## 4. Experiments

In this section, we evaluate the proposed approach on estimating grass biomass and identifying fire-prone road regions. We also show the robustness of the proposed approach and compare its performance with existing approaches.

### *4.1 Experimental Data*

There is no public dataset available that can be used for evaluating the performance of the proposed approach. We collected our own data from a total of 61 roadside sites (named F001 to F061) along the state roads within the Fitzroy region, Queensland, Australia. The ground truths of objective biomass were obtained by marking one square meter area in each of these sites using a quadrat, cutting and bagging the above-ground grass samples in the area, and then storing the samples in a heater (70 °C) for drying up for more than 72 hours, and finally weighting over-dry samples to calculate their fuel loads (tonnes/ha) using a standard formula.

Except for objective biomass, an image of 1936×1296 pixels was also captured for each site using a Dikon D80 camera. To facilitate direct comparisons of the estimated results with ground truth biomass, all images were taken by forcing the following constraints to camera settings: a) the lens directly faces roadside grasses, b) having the same height across all sites, and c) having the same distance from the camera to the sampling region for all sites. The strict settings for the camera are to ensure the accuracy of experimental results. In practice, it is nearly impossible to enforce the same settings to all road sites. Fortunately, the real-world test data for the proposed approach was collected by the DTMR using a vehicle-mounted camera, which was set to keep the same distance from the road boundary as much as possible. Thus, the camera has the same height for all road sites and we can assume that the camera has the same distances and angles to roadside grasses in the same relative locations across the captured images. For example, grass directly in front of the camera is at the same distance and angle for all images. For experiments in this paper, the image regions, which correspond exactly to the sampling regions for all sites, are manually cropped and used as the input to the proposed approach. To provide ground truths of grass density, all image regions are classified into one of three categories: sparse, moderate or dense, based on human visual observation. Samples of the three categories are shown in Fig. 4. The objective biomass and density categories of all images are listed in Table 1.

*4.2 System Parameters*

The ANN used for classifying grass vs. non-grass pixels has a structure of 23-$N$-2 neurons where $N$ is the number of hidden neurons and is set to 16 based on experimental comparisons. The ANN is trained using the Levenberg-Marquardt backpropagation algorithm with a goal error of 0.001 and a maximum epoch of 200. The training data comprises of 650 manually cropped grass and non-grass regions (e.g., tree, road, sky and soil), which are available at https://sites.google.com/site/cqucins/projects. The Gaussian filters for visual feature extraction have a kernel of 7×7 pixels, while the Gabor filters have a kernel size of 11×11 pixels, four orientations $\theta$ = ($0^o$, $45^o$, $90^o$, and $135^o$) and five scales ($\phi_m = f_{max}/(\sqrt{2})^m, m = 0,1,\ldots,4$ and $f_{max} = 0.25$). The width $w_j$ of the surrounding region of each column is set to be 5.

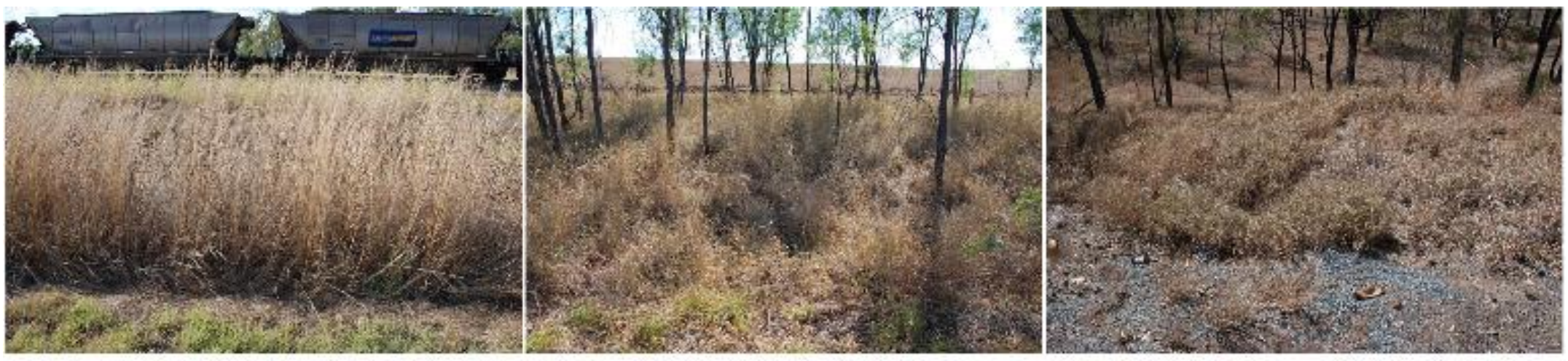

**Fig. 4.** Samples of dense, moderate and sparse grasses.

**Table** 1
Biomass Ground Truths and Estimated DWCGP for All Samples (Unit: Tonnes/Ha).

| Sparse | | | Moderate | | | Dense | | |
|---|---|---|---|---|---|---|---|---|
| No | Biomass | DWCGP | No | Biomass | DWCGP | No | Biomass | DWCGP |
| 07 | 7.94 | 25.2 | 02 | 8.31 | 30.4 | 01 | 23.80 | 35.8 |
| 08 | 6.10 | 37 | 04 | 5.00 | 30.2 | 03 | 28.27 | 27.4 |
| 14 | 15.46 | 22.4 | 06 | 10.68 | 24.1 | 05 | 20.57 | 47.4 |
| 17 | 4.28 | 14.2 | 11 | 0.0 | 15.3 | 09 | 11.74 | 26.6 |
| 19 | 11.60 | 22.4 | 13 | 11.93 | 19.9 | 10 | 16.01 | 44.6 |
| 22 | 6.78 | 23.5 | 18 | 15.87 | 40.1 | 12 | 20.10 | 15.9 |
| 24 | 9.03 | 37.3 | 20 | 14.74 | 32.5 | 15 | 32.10 | 68.8 |
| 26 | 4.20 | 24.1 | 23 | 23.95 | 26.7 | 16 | 11.46 | 40 |
| 30 | 4.05 | 20.1 | 27 | 7.12 | 31.2 | 21 | 11.95 | 39.7 |
| 33 | 11.75 | 17.1 | 29 | 13.60 | 21.8 | 25 | 10.96 | 42.7 |
| 34 | 2.45 | 23.4 | 32 | 13.50 | 31.2 | 28 | 21.24 | 35.3 |
| 35 | 4.15 | 28.4 | 36 | 10.85 | 32.2 | 31 | 13.15 | 35.7 |
| 38 | 18.90 | 22.9 | 40 | 14.85 | 37.1 | 37 | 16.00 | 55.3 |
| 39 | 10.90 | 20.6 | 41 | 20.10 | 36.1 | 44 | 7.20 | 15.1 |
| 42 | 14.55 | 25.1 | 45 | 5.50 | 23.4 | 46 | 14.85 | 31.5 |
| 43 | 3.45 | 18.9 | 47 | 10.25 | 8.8 | 53 | 10.35 | 22.7 |
| 48 | 5.50 | 5.2 | 49 | 11.45 | 35.6 | 54 | 22.85 | 35.2 |
| 50 | 6.85 | 2.3 | 52 | 8.30 | 20.4 | 57 | 17.15 | 47.8 |
| 51 | 2.20 | 12.5 | 55 | 13.10 | 27.5 | 59 | 12.20 | 19.5 |
| 56 | 6.95 | 18.7 | 61 | 8.15 | 16.7 | - | - | - |
| 58 | 7.90 | 13.3 | - | - | - | - | - | - |
| 60 | 4.15 | 3.2 | - | - | - | - | - | - |
| Mean | 7.7 | 19.9 | Mean | 11.4 | 27.1 | Mean | 16.9 | 36.2 |

### 4.3 *Performance of Brown Grass Segmentation*

An important step in the proposed approach is to classify brown grass vs. non-brown grass pixels from roadside images. The classified regions are further used for calculating DWCGP and thus they may have big impact on the prediction results. This part shows the performance of brown grass segmentation. The evaluation dataset includes 50 frames that were randomly selected from the video data collected by the DTMR, Queensland, Australia. The frames cover the most common roadside objects such as brown grass, green grass, tree, soil, road, and sky, as well as various realistic environmental conditions. The pixel-wise ground truths of brown vs. non-brown grasses were manually annotated for all images. The category of non-brown grasses includes all objects other than brown grasses, such as tree, soil, road, and sky. Table 2 shows the classification accuracy obtained using the ANN classifier as compared to an SVM classifier with a RBF kernel. The same set of color and 17-D filter bank based features is used. It can be seen that using ANN slightly outperforms using SVM for both training and test data. Table 3 presents the confusion matrix using the ANN classifier. Brown grass pixels are easier for classification than other object pixels, which is expected as objects in the category of non-brown grasses may have big variations in their visual appearance and structure. Fig. 5 displays the segmentation results on several images, and we can observe that soil pixels are prone to be misclassified as brown grass pixels, likely due to their similar color.

**Table 2**
Classification Accuracy (%) of Brown Grass vs. Non-Brown Grass Pixels.

| | ANN | SVM |
|---|---|---|
| Train data | 91.2 | 85.4 |
| Test data | 75.8 | 75.1 |

**Table 3**
Confusion Matrix (%) for Brown Grass vs. Non-Brown Grass Pixels Using an ANN Classifier.

| | | Target Class | |
|---|---|---|---|
| | | Brown Grass | Non-Brown Grass |
| Estimated Class | Brown Grass | 78.0 | 22.0 |
| | Non-Brown Grass | 26.9 | 73.1 |

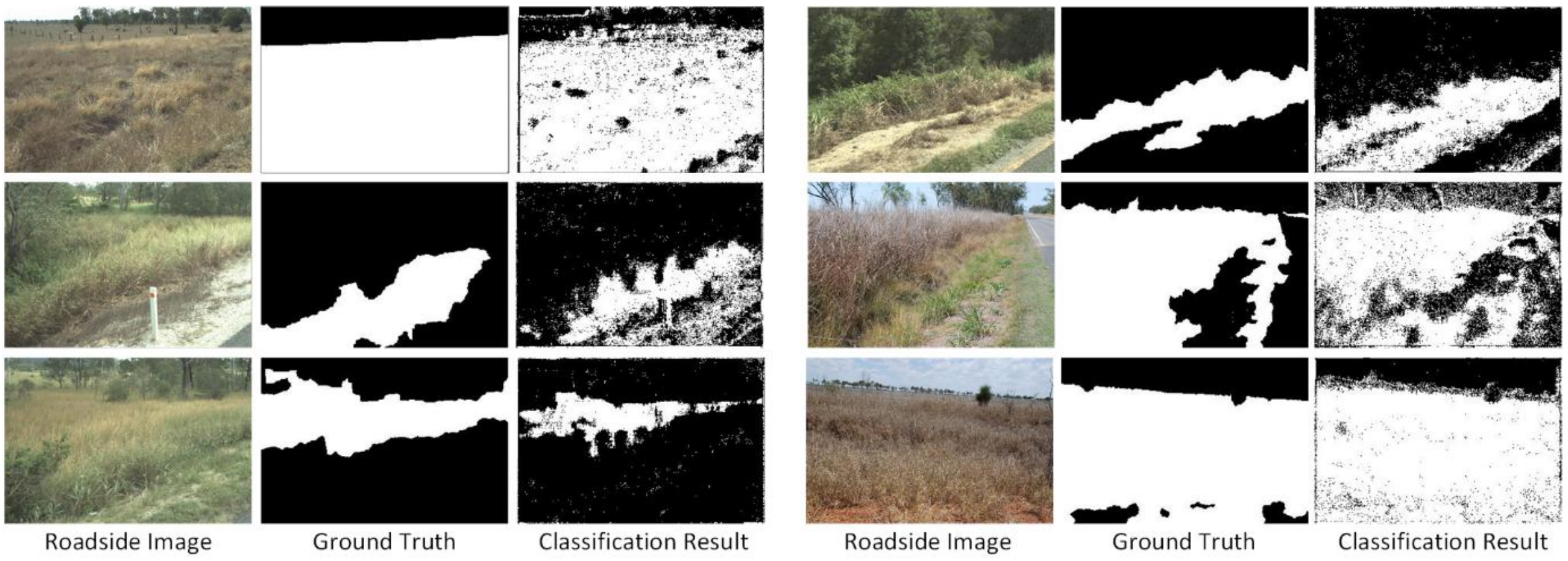

**Fig. 5.** Results of brown grass (white pixel) and non-brown grass (black pixel) classification in sample images.

*4.4 Performance of Biomass Estimation*

The performance of estimating grass biomass is evaluated using Root-Mean-Square Error (RMSE) between DWCGP and objective biomass. The RMSE is a common measure for the performance of biomass estimation in existing work (Clark, Roberts, Ewel, & Clark, 2011), (Anderson, et al., 2018) and it is calculated using $RMSE = \sqrt{\frac{1}{N_I}\sum_{t=1}^{N_I}(D^t - B^t)}$, where $D^t$ and $B^t$ are calculated DWCGP and objective biomass respectively for the $t^{th}$ image and $N_I = 61$ is the number of all images. The calibration factor $a$ in (25) is the quotient of the total biomass divided by the total DWCGP of all samples. The unit of RMSE is the same as biomass unit (i.e., tonnes/ha).

(1) Performance vs. grass density. Table 1 lists estimated DWCGP and the corresponding ground truth biomass for all image samples, as well as their mean values in three density categories. In overall, both DWCGP and biomass tend to have relatively low values for sparse grasses, relatively high values for dense grasses, and medium values for moderate grasses. However, there are also exceptions. For instance, the sample No 38 has comparatively high biomass and DWCGP although it is classified as sparse grasses based human visual observation. By contrast, the sample No 44 has relatively low biomass and DWCGP, but it is classified as dense grasses. The results suggest that, in the categorisation of grass density, even human eyes may make inconsistent predictions with ground truth biomass for individual samples. Thus, it is important to create a reasonably large number of samples to alleviate possible deviations in individual samples.

(2) Performance vs. non-vertical grass stems. In real-world situations, grass stems may grow in various directions and the assumption of a vertical direction in the proposed approach may not be always true. Thus, we also report the results of the proposed approach using image samples with non-vertical grass stems, which are simulated by rotating images by a certain degree. Table 4 shows the RMSEs between DWCGP and biomass obtained using images rotated by [-10$^o$, -5$^o$, 0$^o$, 5$^o$, 10$^o$]. As anticipated, the original images have slightly lower RMSEs than rotated images and the RMSEs tend to increase gradually along with an increased degree. The relatively small differences in RMSEs between original and rotated images is primarily contributed to the invariance of Gabor filter responses to small stem rotations and the adoption of a voting strategy over pre-defined orientations to determine the strongest texture orientation at each pixel. In this paper, Gabor filters categorize the texture responses into four pre-defined orientations (i.e., 0$^o$, 45$^o$, 90$^o$ and 135$^o$) and find the dominant orientation at every pixel by performing a vote on all responses. Thus, texture with small deviations from its dominant orientation due to rotations would still be classified as having the same dominant orientation. For instance, grass stems with an orientation of 80$^o$ will be classified as with a vertical orientation because they are closer to 90$^o$ compared with other pre-defined orientations. The results confirm that the proposed approach is robust to grass stems that are marginally deviated from a vertical orientation.

**Table 4**
Performance of DWCGP Using Rotated Images (RMSE Unit: Tonnes/Ha).

| | Rotation Degree | | | | |
|---|---|---|---|---|---|
| | -10 | -5 | 0 | 5 | 10 |
| RMSE | 5.87 | 5.79 | 5.52 | 5.72 | 5.94 |

(3) Performance vs. Gabor parameters. The DWCGP is designed based on a majority vote over Gabor responses and thus its values may be significantly impacted by Gabor parameters. Table 5 compares the results of a maximum vs. an average operator for producing a single Gabor response for each orientation in (13); and a longest vs. a sum length of continuously connected grass pixels in (15). The two types of parameters directly determine the dominant orientation at each image pixel and DWCGP values respectively. We can observe that: 1) using a maximum Gabor response has slightly lower RMSEs than using an average response, and this probably due to a better capacity of using the maximum value in handling noise across different scales of Gabor filter responses. 2) The use of a longest length of grass pixels has produced lower RMSEs than the use of a sum length, which is expected because a sum length is prone to be affected by short lengths caused by noise in the environment and error in grass region segmentation and dominant orientation detection.

**Table 5**
Performance Comparisons of Maximum vs. average Gabor Responses across Scales, and Longest vs. Sum Lengths of Grass Pixels (RMSE Unit: Tonnes/Ha).

| Gabor Response | Average | | Max | |
|---|---|---|---|---|
| Grass Pixel Length | Longest | Sum | Longest | Sum |
| RMSE | 5.53 | 5.90 | 5.52 | 5.71 |

(4) Performance vs. width of surrounding regions. One key parameter of calculating grass density is the width of surrounding regions for each image column, i.e., $w_j$ in (21). Table 6 shows the results obtained using different values of $w_j$ and the results indicate that using a width of 5 pixels seems to perform the best among the values tested. However, there are only small differences in RMSEs when different widths are used, and thus the width has limited impact on the results.

**Table 6**
RMSE Results vs. Width of Surrounding Regions (RMSE Unit: Tonnes/Ha).

| | 1 | 3 | 5 | 7 | 9 |
|---|---|---|---|---|---|
| RMSE | 5.62 | 5.56 | 5.52 | 5.53 | 5.54 |

(5) Proposed approach vs. previous approaches and human observation. Table 7 shows comparisons of the proposed DWCGP approach with our previous VOCGP approach in (Verma, et al., 2017) and human observation. The VOCGP uses the same set of color and texture features and the same ANN classifier as the proposed DWCGP for brown grass segmentation. Different from DWCGP, it considers only the height of grass stems. The result of human observation is obtained by simply treating the biomass of each image as the mean biomass of its density category (i.e., dense, moderate or sparse). The proposed DWCGP approach shows a lower RMSE than the VOCGP approach and a RMSE close to human observation. The results indicate the possibility of using image processing techniques to achieve biomass estimation results as accurately as humans.

To show the significance of the performance of the proposed approach over benchmark approaches, we conduct statistical tests on the predicted results of all approaches using the two-sample Kolmogorov-Smirnov (KS) test. The KS test is one of the most useful nonparametric methods for comparing the distributions of two data samples. In this paper, the null hypothesis is that the predicted biomass and ground truth biomass come from the same continuous distribution. Observed from Table 7, both the proposed DWCGP and the VOCGP have an *H* of zero, which indicates that both tests fail to reject the null hypothesis and thus the predicted biomass and ground truth biomass come from the same distribution. The DWCGP has a higher *p* value than the VOCGP, indicating its higher accuracy in predicting the biomass. As for human observation, an *H* equal to 1 and a *p* close to zero are obtained. This is expected because the predicted biomass of each image is simply taken as the mean biomass of its density category and thus there are less variations on the distribution of all samples.

**Table 7**
RMSE and KS Test Results of the Proposed DWCGP Approach Compared with the VOCGP Approach and Human Observation (RMSE Unit: Tonnes/Ha).

| | DWCGP | VOCGP | Human Observation |
|---|---|---|---|
| RMSE | 5.52 | 5.84 | 5.49 |
| H | 0 | 0 | 1 |
| p | 0.4877 | 0.1086 | 0.0002 |

Note: the significance level for the KS test is set to be 0.05. If H=1, the null hypothesis is rejected. If H=0, the null hypothesis is not rejected. $p \in [0,1]$ indicates the probability of observing a test statistic as extreme as, or more extreme than, the observed value under the null hypothesis.

(6) Proposed approach vs. benchmark approaches. The current literature still lacks approaches to grass biomass estimation that can be directly included for performance comparisons with the proposed approach. We compare the proposed approach with eight benchmark approaches, which use four texture feature descriptors and two kernel based prediction algorithms, as shown in Table 8. The feature descriptors are selected because they are the most widely used features to represent visual characteristics of objects in various computer vision tasks and have achieved state-of-the-art performance (Ahmed, Rasool, Afzal, & Siddiqi, 2017),(Soltanpour, Boufama, & Jonathan Wu, 2017). Similarly, the SVR and Kernel Ridge Regression (KRR) are also two popular prediction algorithms for regression (Huang, Han, & De la Torre, 2017). Thus, it is anticipated that they are qualified to be used as benchmark approaches to represent the state-of-the-art performance of existing machine learning solutions for automatic biomass prediction. The RMSEs of benchmark approaches are calculated based on five-fold cross validations. In each validation, sampling images from four folds are used for training and those from the rest fold for test. By comparing the results in Tables 7 and 8, we can see that all benchmark approaches have a similar performance of RMSEs around 6.5, except for the approach using Histogram of Oriented Gradients (HOGs) and KRR. The results represent the performance of using state-of-the-art algorithms for grass biomass prediction. The proposed approach has a lower RMSE than all benchmark approaches, showing the effectiveness and feasibility of the proposed concept that utilizes the density weighted connectivity of grass pixels along a vertical orientation to predict grass biomass. The proposed approach outperforms all benchmarked approaches in the KS tests, as the null hypothesis is rejected for all benchmarked approaches. The low performance of benchmark approaches maybe partially due to a limited number of training samples, as collecting large sampling data with objective biomass is time-consuming and effort-intensive. The proposed DWCGP does not suffer from this issue, as its calculation is based on unsupervised learning, which does not require the availability of training data.

**Table 8**
RMSE and KS Test Results of Eight Benchmark Approaches (RMSE Unit: Tonnes/Ha).

| | | | SVR | | | KRR | | |
|---|---|---|---|---|---|---|---|---|
| | LBP | HOG | GLCM | AlexNet | LBP | HOG | GLCM | AlexNet |
| RMSE | 6.26 | 6.49 | 6.72 | 6.54 | 6.37 | 11.70 | 6.56 | 7.39 |
| H | 1 | 1 | 1 | 1 | 1 | 1 | 1 | 1 |
| p | 0.0048 | 0.0000 | 0.0002 | 0.0000 | 0.0002 | 0.0000 | 0.0000 | 0.0005 |

Note: SVR – Support Vector Regression, KRR – Kernel Ridge Regression, LBP – Local Binary Patterns, HOG - Histogram of Oriented Gradients, GLCM - Gray Level Co-occurrence Matrix. AlextNet indicates features learnt from the pre-trained AlexNet (Krizhevsky, Sutskever, & Hinton, 2012) on 61 sampling windows which are resized to 227×277 pixels. A RBF kernel is used for both SVR and KRR.

### *4.5 Performance of High vs. Low Fire Risk Classification*

We also test the performance of low vs. high fire risk classification using the biomass predicted by the proposed approach. For this purpose, we manually selected a total of 382 frames (170 for high and 212 for low fire risk) from the video collected by the DTMR. To simulate real-world situations, no restriction was imposed on the selection process except that the selected frames should contain grass regions of low or high fire risk. In each frame, 15 overlapped sampling regions as shown in Fig. 6 were selected and manually annotated into a category of low, high, or unknown risk. Finally, we obtained 1,298 and 1,641 sampling regions for high and low fire risk respectively. Since the predicted biomass are continuous values, we use a threshold to classify them into a binary category of low or high risk. The threshold is set experimentally to 26.

Table 9 presents the confusion matrix of the classification results. An overall classification accuracy of 87.7% is obtained for all regions. It seems that high risk regions are slightly easier for classification than low risk regions using the proposed approach. Fig. 6 visually displays both good and bad classification results on sample images. In overall, the proposed approach accurately predicts the fire risk levels in most regions. However, as shown in the images in the bottom row, the performance also tends to be impacted by various factors, such as a large rotation degree of grass stems, an excessively dark color in brown grass regions, misclassification of brown grass pixels as soil, and similar texture structure between low and high grasses. These factors represent typical real-world challenging issues that should be paid special attention in further improvements to the proposed approach.

**Table 9**
Confusion Matrix (%) for Low vs. High Fire Risk Classification.

| | | Target Class | |
|---|---|---|---|
| | | High | Low |
| Estimated Class | High | 89.7 | 10.3 |
| | Low | 14.1 | 85.9 |

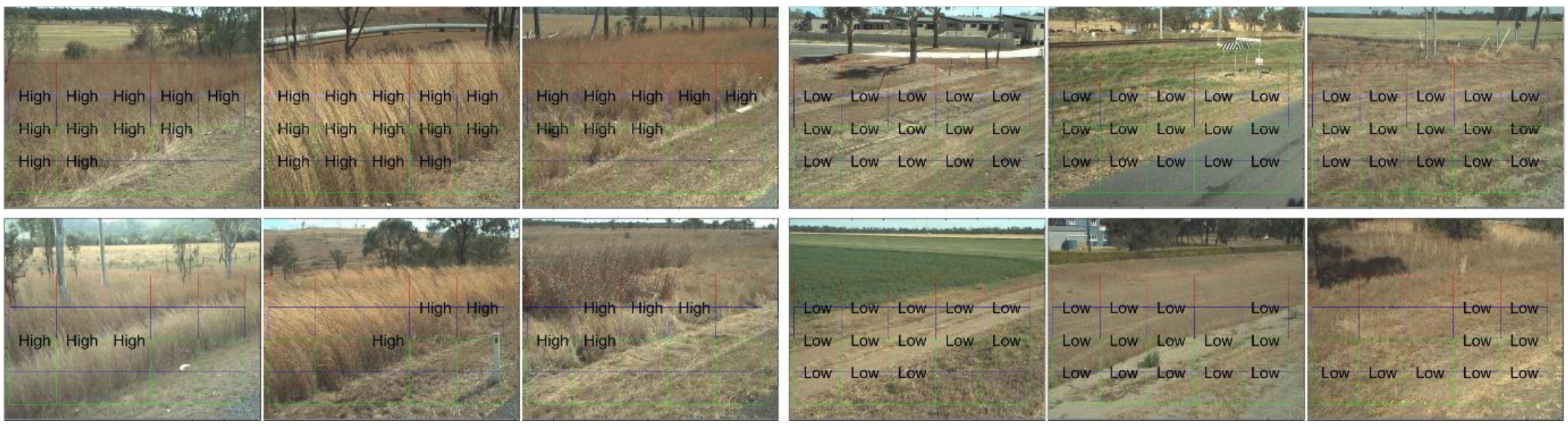

**Fig. 6.** Results of high vs. low fire risk classification using the proposed DWCGP approach in roadsie images. The top row shows images with good classification results, wherease the bottom row displays images with some misclassified regions.

## *4.6 Application to Fire-Prone Road Identification*

One direct application of the proposed approach is to identify fire-prone roadside segments based on the estimated grass biomass. We evaluate the proposed approach on roadside video data collected by the DTMR, Queensland, Australia. A state road No. 16A in the Fitzroy region was chosen for the evaluation and a total of 100 frames were selected from 22 videos that were collected from this road. The frames cover the whole road with approximately the same distance of 200 meters between two frames.

Unlike field surveys where the sampling region is pre-known in every image, there is no information about the locations of sampling regions in the testing video frames. To provide an indication of the biomass in a whole frame, we choose 15 equal and overlapped sampling windows in each frame, as shown in Fig. 7, and obtain an average DWCGP over 15 windows. It is expected that, in images with dense and high grasses, most windows have high predicted DWCGPs and thus a high value of average DWCGP and a high possibility of fire risk. Fig. 8 shows the average DWCGPs of 100 frames that are displayed in an ascending order of their chainage on the road. Typical frames with locally high, low or medium DWCGPs are shown as well. The proposed approach using average DWCGPs achieves very encouraging results since comparisons between locally highest or lowest DWCGPs with the corresponding grass density levels indicate a high level of accuracy. The road segments that have frames of average DWCGPs above a certain threshold (i.e., 26) can be determined as fire-prone regions.

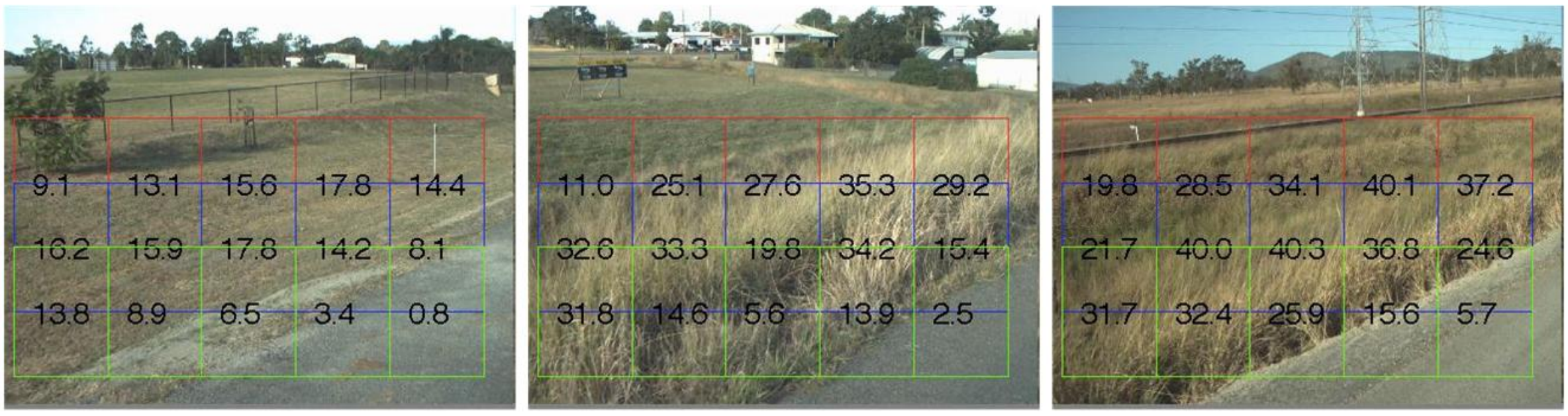

**Fig. 7.** Calculated DWCGPs of 15 grass windows in video frames collected from a state road No. 16A in Fitzroy, Queensland, Australia.

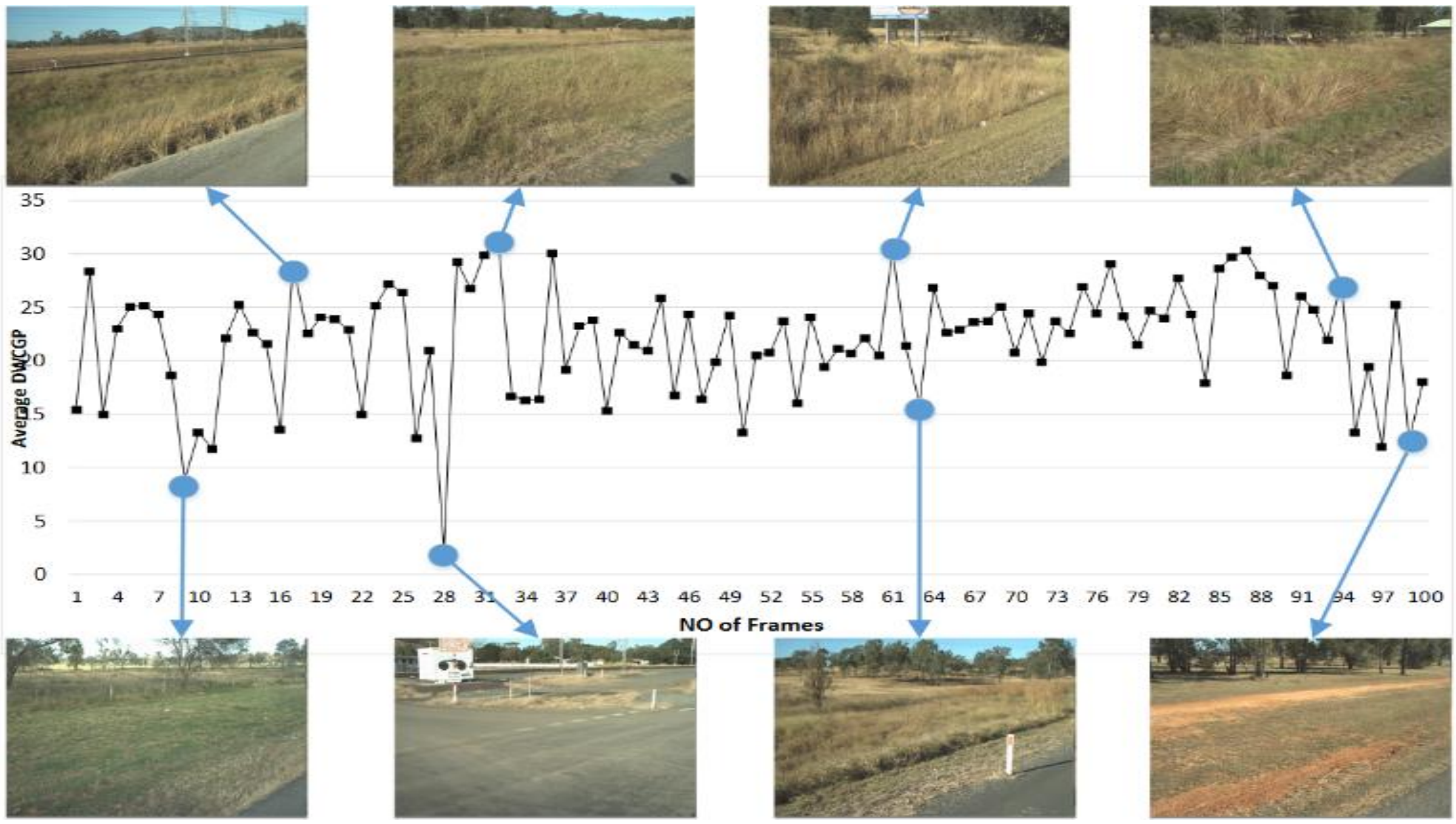

**Fig. 8.** Predicted average DWCGPs in video frames collected from a state road No. 16A. Typical frames corresponding to local maximum or minimum average DWCGPs are displayed. The frames are listed in an ascending order of their chainage on the road.

## 5. Conclusion

Automatically estimating roadside grass biomass from ground-based image data remains largely unexplored in current studies, but it plays a significant role in many practical applications. This paper presented a novel Density Weighted Connectivity of Grass Pixels (DWCGP) approach for the estimation of roadside grass biomass in ground-based images. We conducted extensive experiments on an image dataset to evaluate the effectiveness of the proposed approach in estimating grass biomass. The results show that, compared with the approach that does not consider grass density, the proposed DWCGP reduces RMSE from 5.84 to 5.52. It also has RMSE close to human observation and lower than eight baseline approaches, which use four popular texture features and two kernel regression algorithms. The approach shows good robustness to non-vertical grass stems and is little impacted by using different Gabor parameters and different widths of surrounding regions for calculating grass density. It also demonstrates encouraging results on automatically classifying low vs. high fire risk in image regions and identifying fire-prone road segments.

To further improve the performance of the proposed DWCGP approach in real-life applications, our results indicate that specific attention should be paid to situations such as large rotations of grass stems, and similarity in appearance between brown grasses and other objects. Possible solutions to these issues including incorporating more robust grass segmentation techniques, such as ensembles of prediction algorithms, and considering grass stems in multiple directions (e.g., 45 and 135 degrees) rather than purely a vertical direction during height calculation. Automatic biomass prediction using image processing techniques can be potentially impacted by data collection settings such as the angle and distance of the camera to the vegetation, and the relative location, size, and shape of sampling region within the region. For instance, the same image region captured using different angles or distances of camera may lead to different perception of the grass height, and they are likely to generate different results using the proposed approach. Thus, it is advisable to enforce the same camera settings for all field sites during data collection, however, this is still a difficult issue in real-world practice. It is also beneficial to collect more image data with ground truths of both objective biomass and grass height to facilitate the prediction using supervised learning algorithms.

Except for fire risk assessment, the proposed approach can also be potentially applied to other applications such as vegetation growth condition monitoring, effective vegetation management, and tree regrowth control. Being able to obtain site-specific parameters of roadside vegetation such as biomass, height, coverage and density can provide reliable and important indications of their current condition, growth stage and future tendency. Tacking the changes in these parameters is an effective way to improve effective vegetation management by detecting, quantifying, and handling the possible effects on the vegetation such as diseases, dryness, soil nutrients and water stress. Tree regrowth control aims to reduce road safety threats arising from roadside trees that grow progressively approaching the road boundary. Extending the proposed approach to automatically identifying parameters of these trees such as location, size, species, and distance to the road boundary can support the processes of determining suitable equipment to cut them and accurately predicting the associated cost.

## Acknowledgements

This research was supported under Australian Research Council's Linkage Projects funding scheme (project number LP140100939).